\documentclass{article}

\usepackage{times}
\usepackage{graphicx} % more modern
\usepackage{subfigure} 

\usepackage{natbib}

\usepackage{algorithm}
\usepackage{algorithmic}

\usepackage{hyperref}

\usepackage[accepted]{cs541}

\icmltitlerunning{It Isn’t Sh!tposting, It’s My CAT Posting}

\begin{document} 

\twocolumn[
\icmltitle{It Isn’t Sh!tposting, It’s My CAT Posting}



% you can specify symbols, otherwise they are numbered in order
\icmlsetsymbol{equal}{*}

\begin{icmlauthorlist}
\icmlauthor{Parthsarthi Rawat}{wpi}
\icmlauthor{Sayan Das}{wpi}
\icmlauthor{Jorge Aguirre}{wpi}
\icmlauthor{Akhil Daphara}{wpi}
\end{icmlauthorlist}

\icmlaffiliation{wpi}{Worcester Polytechnic Institute}

\icmlcorrespondingauthor{Parthsarthi Rawat}{prawat@wpi.edu}

\icmlkeywords{Caption Generation, Humor Generation, Nonsensical posts}

\vskip 0.3in
]



\printAffiliationsAndNotice{}  % leave blank if no need to mention equal contribution
%\footnotetext{hi}

\begin{abstract} 
In this paper, we describe a novel architecture which can generate hilarious captions for a given input image. The architecture is split into two halves, i.e. image captioning and hilarious text conversion. The architecture starts with a pretrained CNN model, VGG16 in this implementation, and applies attention LSTM on it to generate normal caption. These normal captions then are fed forward to our hilarious text conversion transformer which converts this text into something hilarious while maintaining the context of the input image. The architecture can also be split into two halves and only the seq2seq transformer can be used to generate hilarious caption by inputting a sentence.This paper aims to help everyday user to be more lazy and hilarious at the same time by generating captions using CATNet.
\end{abstract} 

\section{Introduction}
Comprehending language is challenging enough for machines but producing and detecting humor is considerably more difficult. Teaching computers to perform amusing tasks opens the door to a wide range of practical applications. Streaming platforms can use this to recommend content to suit users’ preferences. It could aid in the translation of wordplays in books and video content, which is a difficult but very useful task to do.

To explore more about this domain, we set ourselves the task of generating a hilarious caption for an image. The biggest hurdle we had was that there wasn’t any structured data to convert images into humorous captions. So, we split our problem into two segments. First, we take an image and generate a normal caption, and then from that generated caption, we convert it into a humorous one using a seq2seq model. While generating humorous captions from normal text data, we also must make sure that we preserve the context of the image.

\subsection{Research contributions}
In this paper, we develop a new model named {\bf CATNet} which aims to generate humorous captions from a given input image. We do not use the traditional image to humor caption pipeline as in \cite{8099591} but break it down into two subsections. The "CA" part of our network take cares of normal captioning using CNN and Attention LSTM while the "T" part of our model converts normal text to hilarious text using seq2seq transformers. {\bf This kind of approach has hardly seen and has shown more hilarious image captioning than most of the models available in literature and offers other unique advantages}.

\section{Related Work}
Advancements in machine translation have provided state-of-the-art captioning methods like \cite{DBLP:journals/corr/ChoMGBSB14} and \cite{NIPS2014_a14ac55a}, using an encoder-decoder framework that outperforms template-based \cite{DBLP:journals/pami/KulkarniPODLCBB13} or search-based \cite{devlin-etal-2015-language} approaches. Our captioning approach i.e. the “CA” part of our CATNet architecture is heavily influenced by \cite{DBLP:journals/corr/XuBKCCSZB15}.\cite{DBLP:journals/corr/DonahueHGRVSD14} and \cite{DBLP:journals/corr/XuBKCCSZB15} introduced CNN as the encoder in the framework to extract the features from the image, and then used RNN as the decoder to generate the descriptions of the image. To improve the relationship between the output sentence and important objects in the image,\cite{DBLP:journals/corr/AndersonHBTJGZ17} and \cite{DBLP:journals/corr/LuXPS16} introduced an attention mechanism in the framework.

Work in humor generation has been limited due to the difficulty in defining humor as well as the difficulty in creating datasets with funny sentences paired with either images or factual sentences. \cite{Ren2017NeuralJG}used LSTM networks to generate funny sentences without any context or with very simple context. The authors in \cite{8099591} created the FlickrStyle10K to generate stylized captions using a CNN encoder and LSTM decoder with a replaceable module for generating captions of a specific style. However, contextual humor generation architectures which do not need paired factual-funny datasets to be trained are still absent. 

\section{Proposed Method}
\label{sec:proposed_method}
In this section, we describe our approach to generate humorous sentences from an input image. It consists of a two-step sequence where we first output a factual caption from an image and then translate that factual sentence into a humorous one.
\subsection{Image Captioning}

 We use VGG-16 as the pretrained CNN model in this paper to extract feature maps. Once these feature maps are extracted attention is applied to it. The attention mechanisms are grouped into two categories: Global\cite{DBLP:journals/corr/LuongPM15} or Local attention\cite{bahdanau2014neural}. The global attention mechanism considers all the hidden states of the encoder when deriving the context vector. This type of model derives a variable-length alignment vector at which uses a scoring function with 3 possible alternatives: dot, general, and concat; we used the general approach. We also implement local attention in our approach which is a combination of hard and soft attention. \cite{DBLP:journals/corr/XuBKCCSZB15} shows “hard” attention to be superior but for our use case we wanted to also include a little bit more background detail of the image.

\begin{table*}[t!]
    \centering
\begin{tabular}{||c c c c c c||}
 \hline
 Appproach & BLEU-1 & BLEU-2 & BLEU-3 & BLEU-4 & METEOR \\ [0.5ex] 
 \hline\hline
 GoogleNIC\cite{DBLP:journals/corr/VinyalsTBE14} & 63 & 41 & 27 & - & - \\ 
 \hline
 Log Bilinear\cite{pmlr-v32-kiros14} & 65.6 & 42.4 & 27.7 & 17.7 & 17.31 \\
 \hline
 SAT(SOTA)\cite{DBLP:journals/corr/XuBKCCSZB15} & 67 & 45.7 & 31.4 & 21.3 & 20.30 \\
 \hline
 {\bf CAT(GLOBAL)} &{\bf 26.42} & {\bf12.38} & {\bf 7.8} & {\bf 3.4} & {\bf 11.54} \\
 \hline
 {\bf CAT(LOCAL)} & {\bf 36.68}  & {\bf 20.1} & {\bf 13.48} & {\bf 6.2} & {\bf 17.39} \\ 
 \hline
 Human & 70 & - & - & - & - \\ [1ex] 
 \hline
\end{tabular}
\caption{BLEU and METEOR metrics comparison of our model to other architectures.}
\label{tab:table1}
\end{table*}

\begin{figure*}[h!]
  \includegraphics[height=5cm]{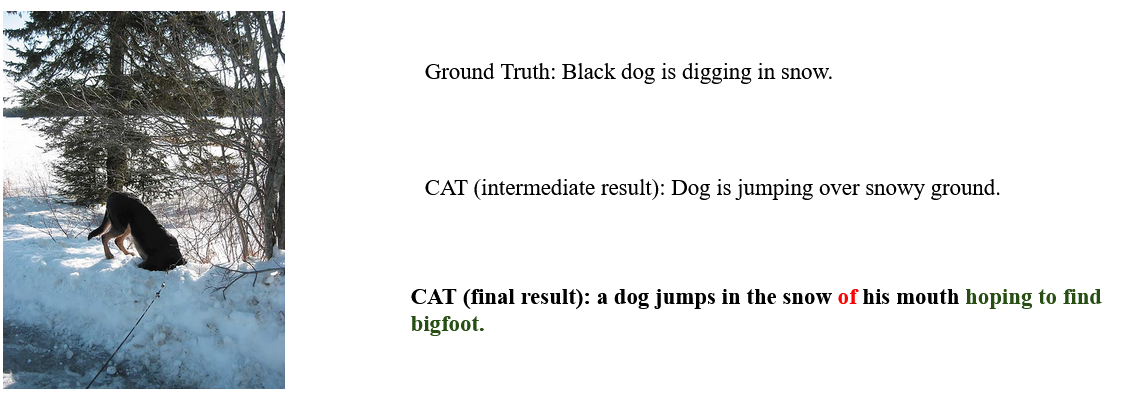}
  \centering
  \caption{Ground Truth,Intermediate and final caption of CATNet}
  \label{fig:fig1}
\end{figure*}

After extracting the feature maps and applying the attention of our choice on these maps we pass the context vector to the decoder RNN which consists of an LSTM model. The LSTM is fed previous predicted word, context vector and hidden states. The context vector is created by calculating a feature mask. This mask is calculated by using previous predicted words and the feature mask. It is then used to parse the feature maps and construct the context vector. The LSTM used as the decoder is aimed to provide enough context to the “T” part of the network to convert the normal generated caption into a humorous one.

\subsection{Humor Generation}
In this section, we describe our method of generating humorous captions from factual ones. The input and output to the model will be two sentences which may be differing in length. Therefore, we use a sequence to sequence model for our task. More specifically, we use a transformer network introduced in \cite{DBLP:journals/corr/VaswaniSPUJGKP17}, which is a seq2seq encoder-decoder network. It differs from previous models in its usage of self-attention layers and other specific components in lieu of recurrent networks. The multi-head attention layer takes in a set of queries, keys and values. Its main aim is to provide a weighted sum of values of one layer to another layer. The weights are based on the relevance of the nodes that provide the values and the ones that receive the final output. If the queries, keys and values come from the same layer, it is called a self-attention layer. Both the encoder and decoder in a transformer comprise several self-attention layers. The encoder applies attention layers on the entire input sequence followed by self-attention layers and fully connected layers. The decoder does the same but takes in the previous words and encoder outputs as inputs. Masking is implemented in the decoder so that it does not take future sequence during training, something which would not be present during inference. This kind of structure makes the network lose the notion of word order in a sentence. To account for this, the transformer adds positional embeddings of input words to its word embeddings. They are representative of the position of words in the sentence and are obtained using sine and cosine functions of different frequencies. This structure enables the transformer to be highly parallelizable as well as retain long term memory.

\section{Experiment}
\subsection{Image Captioning Training Specifics} 
The Image captioning part was trained on Flickr8K dataset which has 8,000 images with 5 captions for every image. We used teacher forcing during training and beam search at the end to generate the best possible caption which can be fed forward to our humor generator.

\subsection{Humour Generation Training Specifics} 

We were hoping to use the FlickrStyle10K dataset for training our humour generation models. It contains images and corresponding triples of captions. Each triplet consists of factual, humorous and romantic/emotional captions. However, we found that only romantic and funny caption pairs were available at the time of training our model, without the corresponding images and factual captions. We initially worked with pairs of emotional and funny sentences as our dataset. Our model was able to generate fun sentences but was still having some issues as the training pairs could contain varying amounts of unrelated information in them. This is illustrated through the following pair - 

{\bf two little girls pose for a sweet picture for their grandmother} 

{\bf two children in orange sunglasses are playing detective } 

To have our model learn better, we had to create a new dataset. On analysing the corpus of funny sentences available to us from the FlickrStyle10K dataset, we realized that most of them consisted of factual sentences with the addition of fun phrases at the end to make them funny. We used this property to generate factual sentences out of the fun sentences themselves. We used tools from Spacy library to extract anchor words like nouns and verbs from the fun sentences and limit the fun sentence to the third last anchor word to generate the factual dataset. This scheme did not always work perfectly but did help improve the performance of our humour model significantly. The following is an example of a factual sentence being generated from a funny sentence. 

{\bf  children sitting in floor playing with several toys and plan to break it }

{\bf three children sitting in floor playing with several toys}

    Due to the small size of our dataset, our model was unable to properly learn the grammar of the language. To remedy this, we decided to first train our model on another dataset to have it learn proper language syntax. Flickr8K is a dataset which contains 5 captions for each image. We created custom training pairs consisting of similar captions from corresponding images of the dataset and pre-trained our model on this dataset. This enabled the network to learn the semantics of this language. The weights were then finetuned for humour generation by training on the fun dataset.

    Our humour generation model was able to output grammatically correct sentences at this point but was still messing up a bit because the training was still not enough to learn proper word representations. For example, it makes sense for a dog to run across fields searching for bones but not for a man to do the same. Although nonsensical outputs such as this can result in humour at times, they were harming output sentence implications in general. To correct this, we implemented a pre-trained word embedding layer in our network so that our model could work with more representative word embeddings. After testing out a few, we finally implemented GloVe embeddings in our network. Teacher Forcing was also implemented during training time.

\begin{figure*}[t!]
  \includegraphics[height=8cm]{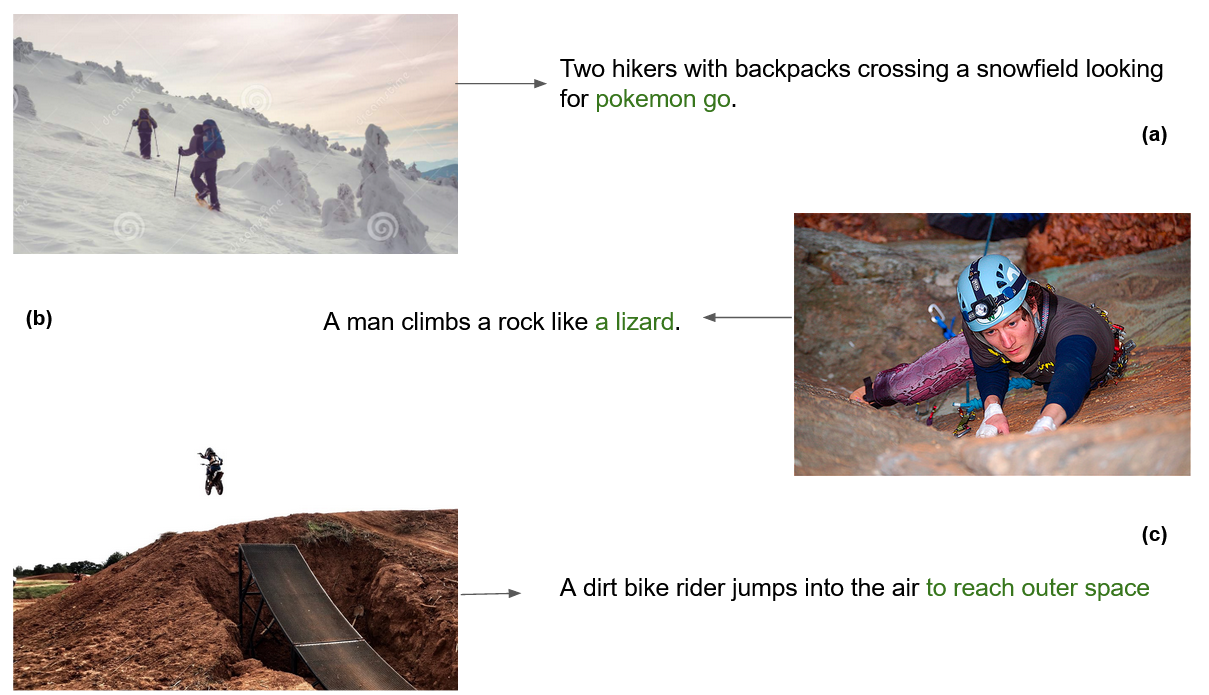}
  \centering
  \caption{A few more generated captions {\bf(a)} Gotta cahtch'em all.... {\bf(b)}Lizard really? {\bf(c)}To the Moon!!!}
  \label{fig:fig2}
\end{figure*}

\begin{figure*}[h!]
  \includegraphics[height=2.5cm]{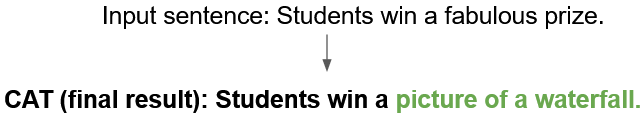}
  \centering
  \caption{Only using the seq2seq part of our model to generate humor.}
  \label{fig:fig3}
\end{figure*}

\section{Results}
We compared our models' BLEU and METEOR scores to the various Image Captioning models available. These metrics are for the first half of the model i.e. the “CA” part because calculating metrics scores for the humor part doesn’t benefit our goal. Usually articles in literature dealing with humor conduct human surveys to provide metrics for their models because of  the same reason.

As shown in Table \ref{tab:table1}. our model has a hard time beating the SOTA and other captioning models which are solely focused on generating as accurate caption as possible. As we are dealing with humor generation as well a comparatively low BLEU score works for our case. Furthermore, we can also see that “local” attention beats “global” attention in every metric. Hence we moved on with the “local” attention as the attention mechanism for our model.

CATNet shows quite impressive results in our trials. An explanation of one such example is shown in Fig. \ref{fig:fig1}. The ground truth for the image was “Black dog is digging in snow.” The captioning part of our model predicts “Dog is jumping over snowy ground.” This is fed forward to the seq2seq part of the model for humor generation which converts it to “a dog jumps in the snow of his mouth hoping to find bigfoot.” Though our model struggles a bit with some grammatical errors it successfully converts the boring caption to a humorous one. A few more comical examples are shown in Fig. \ref{fig:fig2}. Furthermore, we can also split our model and only use the seq2seq part for humour generation using an input sentence. One such example is shown in Fig.\ref{fig:fig3}.

\section{Discussion}
Our architecture often deals with the less descriptive captions. We tried to keep a balance in the description of a caption by keeping the captions descriptive enough to help our humor generation system and not overwhelm it. This lack of description can be improved by extracting better features maps. Complex architectures like ResNet\cite{DBLP:journals/corr/HeZRS15} and DenseNet\cite{DBLP:journals/corr/HuangLW16a} can be used to get richer feature maps. Furthermore, our model was trained on comparatively low data. Flickr8k and FlickrStyle10K was used for training both parts individually. A larger corpus of caption and humor data can help improve the performance of our model. Despite all limitations our model outputs hilarious captions for any input image.

\section{Conclusions and Future Work}
In this paper, we developed a pipeline for generating humorous captions from images. Our modular approach to captioning enables various applications for our framework. Thus, the humor generation module can be used as a standalone system to generate funny versions of normal sentences. A text summarization module can be inserted between the two modules to enable the network to generate one funny caption for a bunch of images. The main obstacle we faced was the lack of proper datasets to train on due to which we had to engineer an approach to create a paired factual dataset from our fun dataset. However, there is a huge corpus of funny text available online.  A future innovation could be the creation of a neural network which can learn the context of humorous sentences themselves and generate new fun sentences when given factual ones. 

\bibliography{paper}
\bibliographystyle{cs541}

\end{document}